\documentclass[conference,a4paper]{ieeetran}
\usepackage[latin1]{inputenc}
\usepackage{subeqnarray}
\usepackage{amsfonts}
\usepackage{amsmath}
\usepackage{amssymb}
\usepackage{graphicx}
\usepackage{color}
\usepackage{float}
\usepackage{float}
\title{A Comparison of CNN-based Face and Head Detectors for Real-Time Video Surveillance Applications}

\author{\authorblockN{Le Thanh Nguyen-Meidine\authorrefmark{1}, Eric Granger \authorrefmark{1}, Madhu Kiran\authorrefmark{1} and Louis-Antoine Blais-Morin\authorrefmark{2}}
\authorblockA{\authorrefmark{1} \'{E}cole de technologie sup\'{e}rieure, Universit\'{e} du Qu\'{e}bec, Montreal, Canada\\ \ 
lethanh@livia.etsmtl.ca, eric.granger@etsmtl.ca, mkiran@livia.etsmtl.ca}
\authorblockA{\authorrefmark{2} Genetec Inc., Montreal, Canada \\ 
lablaismorin@genetec.com}}

\IEEEoverridecommandlockouts

\pubid{\makebox[\columnwidth]{978-1-5386-1842-4/17/\$31.00~\copyright 2017 IEEE \hfill}\hspace{\columnsep}\makebox[\columnwidth]{ }}

\begin{document}
\maketitle

\begin{abstract}
Detecting faces and heads appearing in video feeds are challenging tasks in real-world video surveillance applications due to variations in appearance, occlusions and complex backgrounds. Recently, several CNN architectures have been proposed to increase the accuracy of detectors, although their computational complexity can be an issue, especially for real-time applications, where faces and heads must be detected live using high-resolution cameras. This paper compares the accuracy and complexity of state-of-the-art CNN architectures that are suitable for face and head detection. Single pass and region-based architectures are reviewed and compared empirically to baseline techniques according to accuracy and to time and memory complexity on images from several challenging datasets. The viability of these architectures is analyzed with real-time video surveillance applications in mind. Results suggest that, although CNN  architectures can achieve a very high level of accuracy compared to traditional detectors, their computational cost can represent a limitation for many practical real-time applications. 
\end{abstract}


%%%%%%%%%%%%%%%%%%%%%%%%
\section{Introduction}

The ability to locate and identify multiple individuals at different moments and places is important in many video analytics and surveillance applications, such as scene understanding, action and event recognition, video summarization, person re-identification, and watchlist screening. A person appearing in a camera's field of view is often detected based on the appearance of their face and head, and the corresponding regions of interest can be employed to initiate tracking, analysis and recognition of individuals. Detecting faces and head over a network of video cameras is challenging tasks in real-world surveillance applications due to complex scene and real-time requirements. Implementations must therefore be computationally efficient and scale well to the number, frame rate and resolution of cameras and to the number of persons in the scene.

Face detection has been an active research area in computer vision and pattern recognition for several years due to the challenging nature of face as an object, and to the large number of applications. For instance, the popular cascaded boosting detector proposed by Viola and Jones~\cite{Viola2004} is known to rapidly detect frontal faces with low computational complexity. Multi-view face detectors rely on face descriptors and classifiers that are not typically discriminant enough to detect faces under a wide variety of capture conditions. Indeed, it is often difficult to train such detectors off-line with enough reference images to cover all possible variations in facial appearance (due to changes in illumination, pose, scale, motion blur, etc.) and occlusions that can occur in real-world unconstrained surveillance environments. Moreover, applying a detector to every high resolution video frames captured over a network of cameras in potentially cluttered scenes is a complex task, which limits their real-time application. These issues have driven much research in face detection, and have led to many innovative techniques based on rigid templates, in particular on variations of boosting and neural networks, and on deformable part-based models~\cite{ZAFEIRIOU20151}.

Exploiting the temporal coherence obtained using a visual tracker allows to continuously locate a same person over time, and can improve the accuracy and complexity since they only require local detection of previously-seen faces and heads in a video. However, global detection (on entire frames) would still be required periodically to initiate new person tracks and drop others. In addition, long term tracking of multiple persons in complex scenes based on information extracted from, e.g., faces, heads, bodies, gait, and clothing, is also a challenging problem in real-word video surveillance applications due to appearance changes, deformations, occlusions, and clutter and dynamic backgrounds.  

More recently, several deep learning architectures have been shown to achieve a high level of accuracy in complex visual object (including face and head) detection applications, where robust object  representations are learned directly from large-scale datasets~\cite{Bengio12}. However, the computational complexity of these architectures are known to be high~\cite{SpeedAcc}, especially for real-time applications where faces and heads must be detected live from many high resolution cameras. 

In this paper, state-of-the-art convolutional neural networks (CNNs) suitable for face and head detection are compared according to accuracy and computational complexity. These architectures include include Faster Regional-CNN (R-CNN)~\cite{FasterRCNN}, Region-based Fully Convolutional Networks (R-FCN)~\cite{RFCN}, PVANET~\cite{PVA}, Local R-CNN~\cite{Vu} and Single Shot Multi-Box Detector (SSD)~\cite{SSD}. These CNN architecture are organized according to: (1) single pass approaches (like SSD) that detect with one step through the CNN, and (2) region based approaches (like Faster R-CNN) that exploit a bounding box proposal mechanism prior to detection. They are compared empirically to baseline detectors according to accuracy and time complexity on images from several challenging public datasets. The trade-off of accuracy versus complexity of these architectures are analyzed for real-time video surveillance applications.

%%%%%%%%%%%%%%%%%%%%%%%%%%%%
\section{CNN Architectures for Face Detection}
%%%%%%%%%%%%%%%%%%%%%%%%%%%%

%\begin{figure*}[ht!]
%    \centering
%    \includegraphics[width=17.5cm]{images/VGG16.png}
%    \caption{Architecture of the VGG16 network.}
%    \label{VGG16 Architecture}
%\end{figure*}

This section presents a survey state-of-the-art CNN meta-architectures that are suitable for face and head detection in video surveillance applications. Although each architecture can employ different basic CNNs for accurate feature extraction~\cite{SpeedAcc} -- Inception, ResNet, VGG-16, etc. -- this paper will focus on CNN meta-architectures. They are organized below according to \textit{single pass} and \textit{region based} approaches.

%%%%%%%%%%%%
\subsection{Single Pass CNN Detectors:}

%\subsection{Basic CNN detector:}
%Face and head detection can be performed with a simple architecture consisting of a cascade of convolutional layers and a fully connected layer trained to maximize accuracy on face and head detection data. Figure~\ref{VGG16 Architecture} shows the most basic VGG-16 architecture. As shown in the figure, the input size is fixed which can reduce its versatility in video surveillance. Although other basic CNN architectures are available (like, e.g., ResNet or Inception networks) VGG16 was considered here because it is representative of accurate CNNs for feature extraction.

\begin{figure}[h!]
    \centering
    \includegraphics[width=6.6cm]{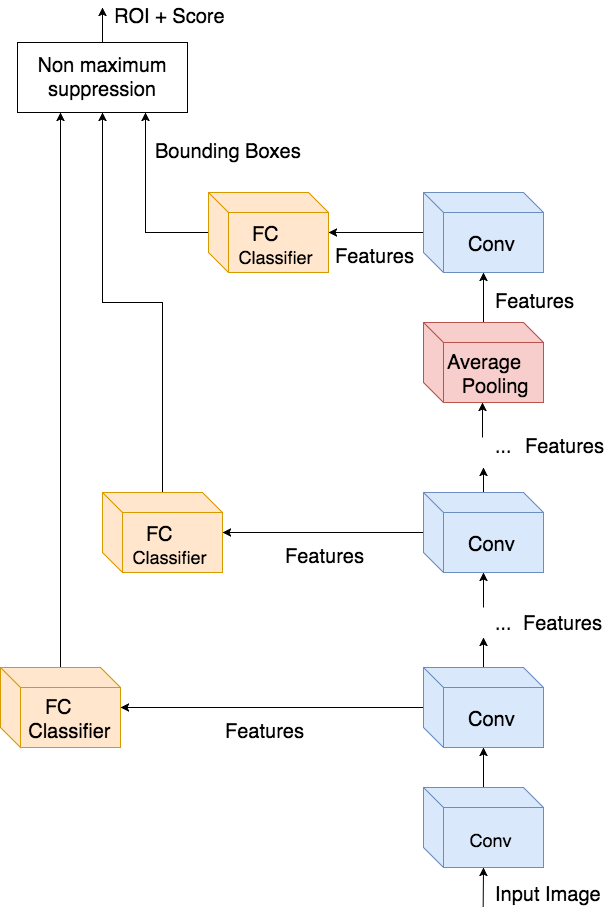}
    \caption{Single Shot Multi-Box Detector (SSD) architecture.}
    \label{SSD Architecture}
\end{figure}

%\subsection{Single Shot Multi-Box:}
Single shot detectors generally employ a single feed-forward convolutional network to directly predict bounding boxes proposals. For the feature extractor, SSD uses VGG-16 as its feature extractor because of its simplicity. Given an input image, the architecture of the Single Shot Multi-Box Detector (SSD)~\cite{SSD} (shown in Figure~\ref{SSD Architecture}) outputs bounding box proposals on some selected convolution layers. At each selected layer, the feature map is divided into a grid of, e.g. 16x16, 8x8, and then each layer computes a score per box. Then, the bounding boxes are processed by the  corresponding fully connected classifier, which further eliminates box proposals. The remaining bounding boxes output from these classifiers are then merged using Non-Maximum Suppression.

%%%%%%%%%%%%%%%%%%%%%%
\subsection{Region-Based CNN Detectors:}

%\subsection{Faster R-CNN~\cite{FasterRCNN}}
Region-based architectures are among the most promising approaches, where bounding box proposal mechanisms are employed for accurate object detection. R-CNN~\cite{RCNN} is one of the first region-based CNN detectors. It uses selective search to generates region proposals, and each proposal requires a forward pass through the whole network to compute features. For example, a Local R-CNN has been exploited by Vu et al.~\cite{Vu} for head detection. Instead of classifying head versus non-head regions with SVMs, the authors employ the output of a pre-trained CNN~\cite{Oquab} to compute candidate scores. R-CNN is often seen as a complex architecture because of the selective search process and forward passes for each proposed region. 

\begin{figure}[!th]
    \centering
    \includegraphics[width=4cm]{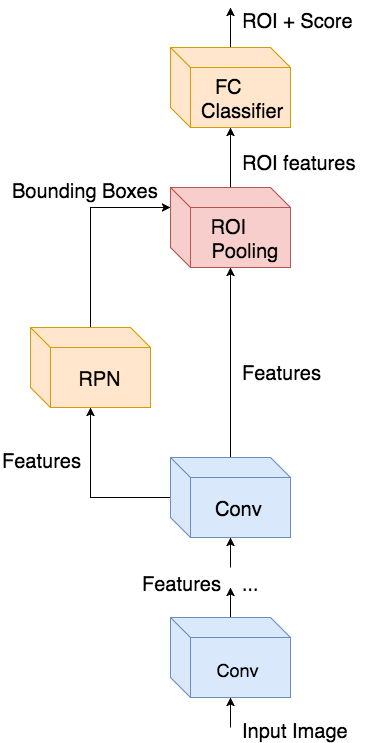}
    \caption{Faster Regional-CNN (R-CNN) architecture.}
    \label{FasterRCNN}
\end{figure}

Currently, Faster R-CNN~\cite{FasterRCNN} (shown in Figure~\ref{FasterRCNN}) is the most popular architecture of this type, and has inspired several other architectures. Instead of using a selective search like the previously-proposed architectures (R-CNN~\cite{RCNN} and Fast R-CNN~\cite{FastRCNN}), this approach exploits a Region Proposal Network (RPN).  The RPN generates many bounding box (ROI) proposals for classification, while the classifier pools ROI features from the shared convolutional layers. 

Given an input image, features output from an intermediate convolutional layer are presented to the RPN, which contains another convolutional layer and a regression layer to predict bounding boxes, and a box classification layer. The ROI pooling layer receives bounding box proposal from the RPN, and features from the previous convolutional layer. The classifier then uses the features pooled by the ROI pooling layer to produce scores, allowing to predict whether or not the ROI is a face/head. An advantage of region-based CNN detectors is that all ROIs are transformed to a fixed size feature map prior classification, allowing for a variable size for input images. However, the processing time depends on the number of boxes proposed by the RPN.

\begin{figure}[!th]
    \centering
    \includegraphics[height=6.4cm]{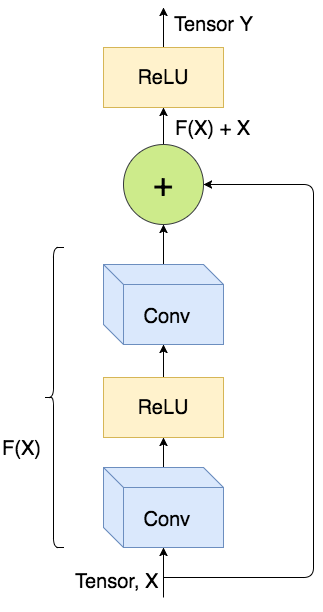}
    \caption{Structure of a residual layer.}
    \label{ResLayer}
\end{figure}

\begin{figure}[th!]
    \centering
    \includegraphics[width=4cm]{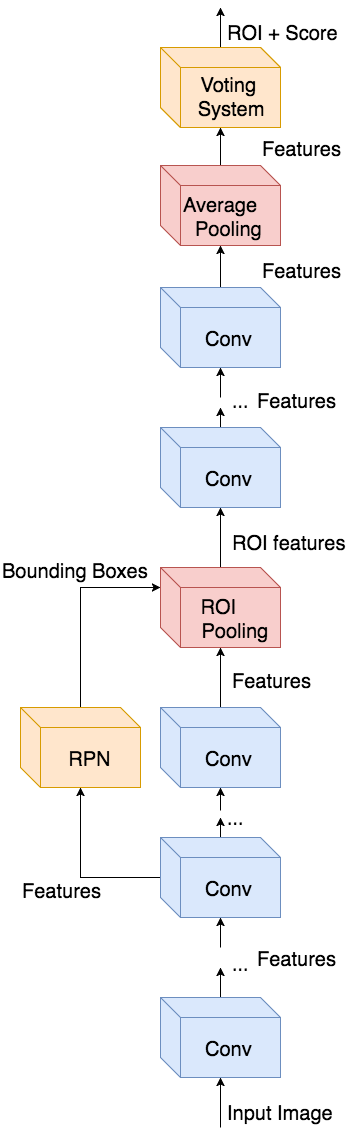}
    \caption{Region-based Fully Convolutional Networks (R-FCN) architecture.}
    \label{RFCN}
\end{figure}

%\subsection{R-FCN~\cite{RFCN}}
Region-based Fully Convolutional Networks (R-FCN)~\cite{RFCN} shown in Figure~\ref{RFCN} is another architecture based on RPNs. It is proposed to improve the efficiency of Faster R-CNN~\cite{FasterRCNN} by using a residual network for feature extraction, and a fully convolutional layer for classification. Residual networks~\cite{ResNet} work on the assumption that, instead of only learning information output from the previous layer, information from the input to this previous layer can also provide valuable information for training. This is particularly true when the residual layer (ReLU) does not activate, and the output from the previous layer is zero. As illustrated in Figure~\ref{ResLayer}, residual learning adds the input of the previous layer to its output, and inputs this sum to the next layer.  

The R-FCN and Faster-RCNN architecture are very similar, except that the RPN in the R-FCN extracts features prior to last layer of the feature extractor, and the architecture uses a fully convolutional network (there is no fully connected layer at the end). Instead, the R-FCN uses a voting system -- the average score of the ROI from the average pooling layer -- to determine whether an ROI is an object of interest. In particular, features are extracted from the last layer prior to prediction instead of extracting features from the same layer where region proposals are predicted. This minimizes the amount of per-region computations. The RPN and classifier share more convolution layers than with Faster R-CNN. Finally, to provide translation-variance representations, R-FCN uses a position-sensitive extraction mechanism instead of standard ROI pooling techniques.

\begin{figure}[h!]
    \centering
    \includegraphics[width=5.2cm]{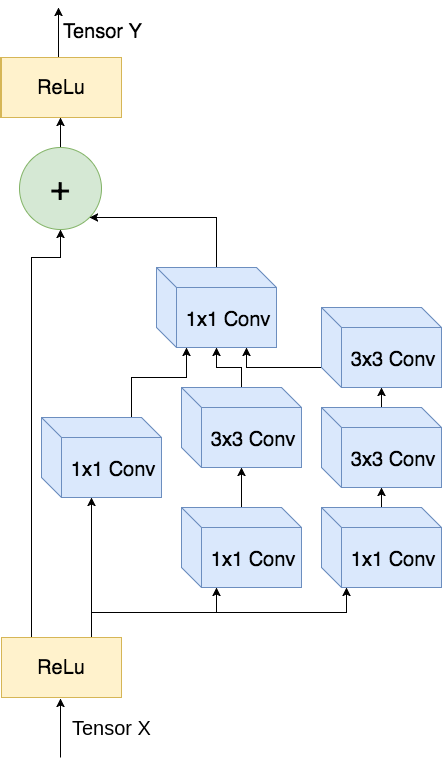}
    \caption{Structure of an Inception-Resnet-v2 layer.}
    \label{InceptionResNetV2}
\end{figure}

%\subsection{PVANet }
Unlike the two previous techniques, PVANET~\cite{PVA} uses an Inception-ResNet network (see Figure \ref{InceptionResNetV2}) for feature extraction. This recently-proposed CNN detector combines the strength of Inception and ResNet models, where the Inception model introduces the notion of parallel feature extraction into its architecture. Instead of choosing the network's filter size at each layer, e.g. 1x1, 3x3 or 5x5, Inception uses all of them, and allows the network to select the best features. This allows the model to learn multi-level features from multiple filters to improve performance. Moreover, combining Inception and ResNet models can potentially achieve higher level of accuracy than with either Inception or ResNet alone. 

\begin{figure}[h!]
    \centering
    \includegraphics[width=7.5cm]{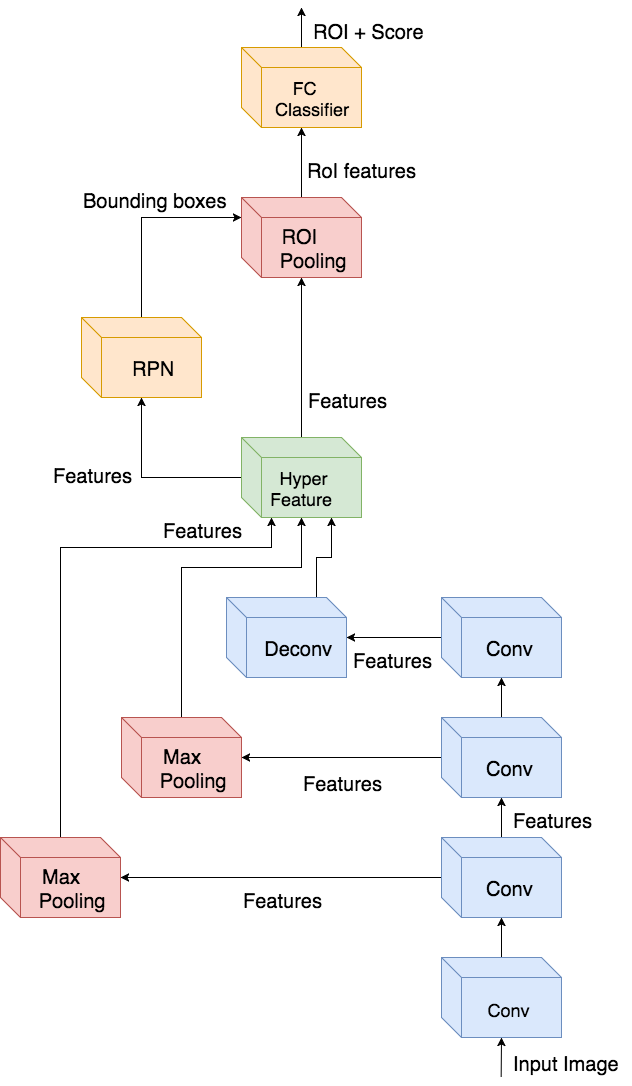}
    \caption{PVANET architecture.}
    \label{PVA}
\end{figure}

The PVANET architecture (see Figure \ref{PVA}) uses Hyper Feature~\cite{HyperNet} to combine features for the RPN and FC classification, instead of using the output of a single feature extraction layer as the input to the RPN. Hyper Feature concatenates the features extracted from lower-levels, and a applies a deconvolution operation to upscale higher-level features. The features map produced by Hyper Feature is also smaller than the one employed in R-FCN and Faster R-CNN, thereby reducing the computational complexity. 

%%%%%%%%%%%%%%%%%
\section{Experimental Methodology}
%%%%%%%%%%%%%%%%%

%\subsection{Protocol}
In order to compare the CNN detectors described in Section 2, each architecture was implemented for GPU processing using Caffe deep learning framework (see Table~\ref{SumMA}). Baseline detectors were implemented with OpenCV for Viola-Jones, and Matlab with C backend for HeadHunter DPM. All the CNN architectures were initially pretrained with images from the PASCAL VOC database for object detection \cite{PascalVoc}.  Then, these architectures were fine-tuned on Wider Face~\cite{WIDER} dataset for face detection experiments. Average test-set performance was calculated on facial images of the FDDB~\cite{FDDB} dataset. For head detection experiments, all the architectures were fine-tuned with Hollywood Head~\cite{Vu} dataset, and then tested on the Casablanca~\cite{VJ-CRF} dataset. These publicly datasets are comprised of many face and head images captured in a wide range of challenging conditions. For all the region-based architectures, images were fed 'as is' since they can handle different image sizes thanks to ROI Pooling. As for SSD, we used the SSD300 model for testing, the input images were reshaped into a 300x300 tensor to fit the network. The rest of this section provides additional details on the dataset and performance metrics used in this paper. 

\begin{table}[b!]
\centering
\caption{Summary of implementations for each CNN architecture.}
	\begin{tabular}{ |c||c|c|c|c|| } 
 	\hline
 	\textbf{Detector}				& \textbf{Feature Extractor}  & \textbf{Matching} 	& \textbf{Loss Function}  		\\ \hline \hline
    SSD~\cite{SSD}             		& VGG-16   			&  argmax &	smooth $L_1$	\\ \hline
    Faster R-CNN~\cite{FasterRCNN}  & VGG-16   			&  argmax &	smooth $L_1$	\\ \hline
    R-FCN~\cite{RFCN}    			& ResNet50/101   	&  argmax & smooth $L_1$	\\ \hline
    PVANET~\cite{PVA}   			& Inception-Resnet  & argmax  & smooth $L_1$	\\ \hline
  \end{tabular}
\label{SumMA}
\end{table}
  
%\subsection{Datasets}
Hollywood Heads is a large-scale dataset~\cite{Vu} derived from 21 Hollywood movies. They include a total of 369,846 human heads annotated in 224,740 color RGB video frames, with various resolutions, e.g., 528x224 and 640x360. The movies vary over various genres and time epochs.  The Casablanca dataset~\cite{VJ-CRF} contains frames exacted from the movie Casablanca. It is comprised a total of 1466 frames with annotated head bounding boxes. The Casablanca dataset is annotated like the Hollywood dataset except that the frontal head annotation have been reduced to faces. The resolution of all the greyscale frame is 976x720p. Casablanca is an archive film and provides an interesting setting for head detection, where images greyscale, with poor lighting and crowded scenes.

Wider Face is a large-scale face detection dataset~\cite{WIDER}. It consists of 32,203 images with 393,703 labeled faces. The resolution of all the images is varied 1024x696 an 1024x1048. The faces have been selected based changes in in appearance based on pose and scale variations. The annotation indicates various attributes such as occlusion, pose and categories. The Face Detection Dataset and Benchmark (FDDB)~\cite{FDDB} dataset is a collection of labeled faces from Faces in the Wild dataset. It contains a total of 5171 face annotations, where images are also of various resolution, e.g. 363x450 and 229x410. The dataset incorporates a range of challenges, including difficult pose angles, out-of-focus faces and low resolution. Both greyscale and color images are included.  

\begin{figure*}[th!]
    \centering
    \includegraphics[width=7.6cm]{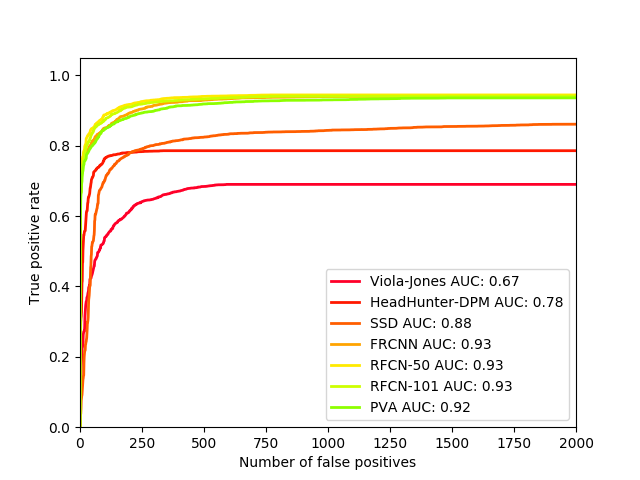}
    \includegraphics[width=7.6cm]{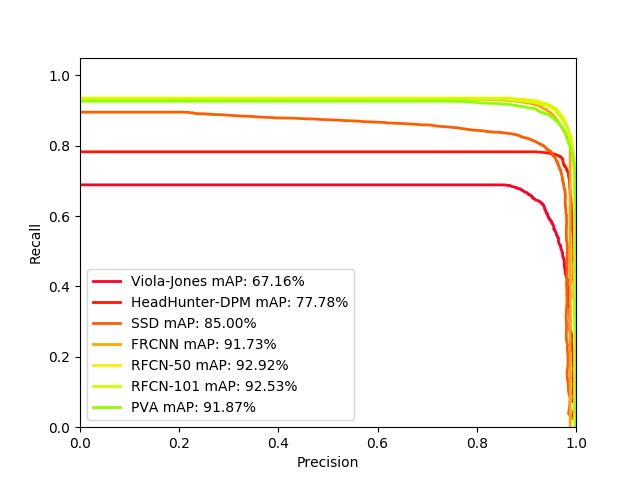}
    \caption{ROC and inverted Precision-Recall curves for face detectors on FDDB data.}
    \label{FDDBPR}
\end{figure*}

\begin{figure*}[th!]
    \centering
    \includegraphics[width=7.6cm]{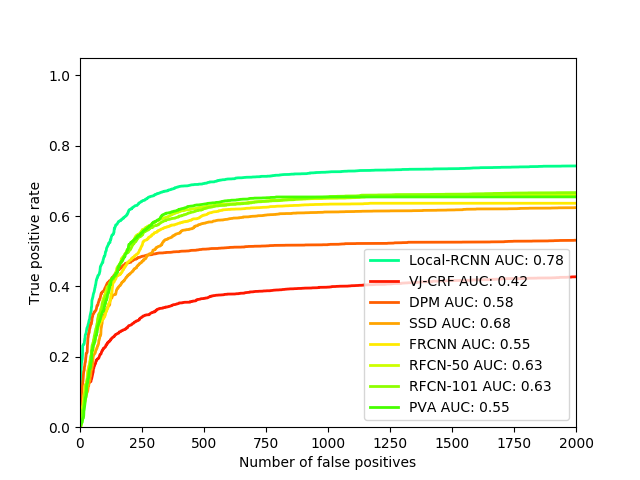}
    \includegraphics[width=7.6cm]{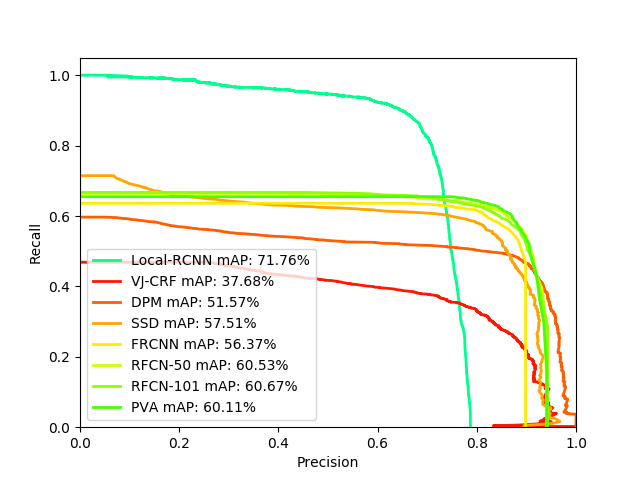}
    \caption{ROC and inverted Precision-Recall curves for head detectors on Casablanca data.}
    \label{CasaPR}
\end{figure*}

%\subsection{Performance metrics}
Different performance metrics were considered to compare the CNN architectures for face and head detection. Accuracy is measured using the ROC curves, in which true positive rate (TPR) is plotted as a function of false positive rate (FPR) over all decision thresholds. TPR is defined as the proportion of target face ROIs that are correctly detected as face over the total number of ground truth ROIs in the sequence. Meanwhile, FPR is the proportion of non-target face and head ROIs incorrectly detected as face over the total number of non-target face and head ROIs. The area under ROC curve (AUC) is a global scalar metric of the detection performance. 
 
ROC curves and AUC measures allow for an evaluation of performance that is independent of miss-classification costs and class priors between classifiers. In video surveillance, the number of faces varies time in each sequence. Thus, the Precision-Recall space is more suitable for measuring detector performance under imbalanced data situation. Recall corresponds to TPR and precision (P) is computed as follows P = TP/(TP + FP). The mean average precision (mAP) measures the average precision of the detector over vertices of the precision-recall curve. It is calculated using: $\mbox{mAP} = \scalebox{1}{$\sum_{k=1}^{n} P(k) \Delta r(k)$}$, where \emph{P(k)} is the precision measure at decision threshold \emph{k}, and \emph{$\Delta$r(k)} as the difference in recall from threshold \emph{k - 1} to \emph{k}. The mAP measures corresponds to the area under the Precision-Recall curve.

\begin{table}[b!]
	\centering
	\caption{Average AUC and mAP accuracy for face and head detection on the FDDB dataset and Casablanca dataset.}
	\label{AccuracyTable}
	\begin{tabular}{|c||c|c||c|c||}
    \hline
  	\textbf{Detector}	& \multicolumn{2}{|c||}{\textbf{FDDB data}}  & \multicolumn{2}{|c||}{\textbf{Casablanca data}}   \\  
    					& AUC 		& mAP (\%) & AUC & mAP (\%)	\\ \hline \hline
  	VJ~\cite{Viola2004}$\backslash$VJ-CRF~\cite{VJ-CRF}     	& 0.6654  	& 67.16  & 0.42 	& 37.68		\\ \hline
    HeadHunter~\cite{DPM}$\backslash$DPM~\cite{originalDPM}		& 0.7780  	& 77.78  & 0.58 	& 51.57 	\\ \hline \hline
    SSD~\cite{SSD}             			& 0.7740  	& 85.00  & 0.68	& 57.57		\\ \hline
    Faster R-CNN~\cite{FasterRCNN}    	& 0.9253 	& 91.73  & 0.55	& 56.37 	\\ \hline
    R-FCN 50~\cite{RFCN}        		& 0.9339  	& 92.92  & 0.63	& 60.53 	\\ \hline
    R-FCN 101~\cite{RFCN}       		& 0.9287  	& 92.53	 & 0.63	& 60.67		\\ \hline
    PVANET~\cite{PVA}    				& 0.9168  	& 91.87  & 0.55	& 60.11 	\\ \hline
    Local-RCNN~\cite{Vu} 				& N/A		& N/A	 & 0.78 & 71.76 	\\ \hline
\end{tabular}
\end{table}

Time complexity was measured as the average number of frames per second (FPS). Since this measure depends on several external factors to the algorithm (e.g., the GPU, CPU, etc.), the number of floating point operations (FLOP) was also considered to measure the runtime performance because it is less dependent of the hardware, although it is still dependent of the software. All experiments were performed using an Intel Xeon CPU E3-1270 (3.60GHz) computer with Nvidia M4000 GPU acceleration card. This type of computer corresponds to a cost-effective hardware solution for many industrial applications. The monitoring and profiling tools provided by NVIDIA (nvprof and nvvp) was used to measure the number of FLOPS and the memory cost of each algorithm.

%%%%%%%%%%%%%%%%%%%%%%%%%%%%%%%%%%
\section{Results and Discussions}
%%%%%%%%%%%%%%%%%%%%%%%%%%%%%%%%%%

%\paragraph{Accuracy}
Figure~\ref{FDDBPR}\footnote{We attempted to compare CNN detectors on ChokePoint and COX Face DB, two public datasets for face recognition in video surveillance \cite{Saman2017}. With these datasets, occluded faces are not annotated. Since the CNN detectors are very accurate and detect several occluded faces without eyes, many false positives were introduced into our evaluation. The results were therefore not included in this paper.} compares the ROC and Precision-Recall curves for traditional and CNN-based face detectors fine-tuned on Wider Face dataset, and then tested on the FDDB dataset. The area under these curves is summarized in Table \ref{AccuracyTable}.  Results show that the CNN architecture provide a significantly higher level of accuracy versus traditional methods like Viola-Jones and HeadHunter DPM. Both R-FCN networks outperform other detectors in terms of ROC accuracy, although the three region-based CNNs -- Faster R-CNN, R-FCN and PVANET -- provide a comparably high level of accuracy that is significantly higher than SSD. However, SSD provides more competitive performance in the Precision-Recall space. The accuracy obtained by using a very recent single-shot CNN architecture called Yolo 9000 \cite{Yolo9000} on our datasets is not competitive so we decided not to present them in this paper.

Figure \ref{CasaPR} compares the traditional and CNN-based detectors for head detection. These detectors were fine-tuned on Hollywood Head Datasets and tested on Casablanca dataset. The results show that compared to the face detection problem, head detection is still very challenging. Among CNN detectors, Local R-CNN achieves the highest level of accuracy. Note that Local R-CNN was considered here because it provided a baseline for head detection,  but its time complexity is too high for video surveillance applications.

Our results also compare ResNet50 and ResNet101 inside the R-FCN architecture. The R-FCN with ResNet50 is slightly more accurate than the ResNet101, but the difference is almost negligible. Results show one of the advantages of R-FCN is its residual learning. Even when reducing the number of R-FCN layers, the network sustains a high level of accuracy since residual layers learn much better than regular convolutional layer. Also, since the RPN and classifier share more convolution layers than with Faster R-CNN, allowing for a more compact FC classifier. Since this classifier will process each ROI proposed by the RPN, this also translated to a faster processing time than Faster R-CNN.

%\paragraph{Complexity}
Table \ref{Time and memory complexity} shows the average time and memory complexity of traditional and CNN-based face detectors on the FDDB dataset. Time complexity was measures on a 450 x 387 pixel images with 10 box proposals. The memory consumption was measured on the same image with a batch size of 1. The SSD and Yolo 9000 detectors represent the more efficient solutions, where a face can be detected in these images at about 19 FPS using relatively compact implementations. Regardless, these results indicate that even with the fastest CNN architectures, the time complexity is high compared to the Viola-Jones detector, but potentially suitable for real-time industrial applications in video surveillance. 
\begin{table}[b!]
\centering
\caption{Average time and memory complexity for face detection on FDDB.}
\begin{tabular}{ |l||c|c|c| } 
 \hline
 	\textbf{Detector}	& \multicolumn{2}{|c|}{\textbf{Time}} & \multicolumn{1}{|c|}{\textbf{Memory}}  \\ %\cline{2-4} 
   						& \textbf{GFLOPS}  	& \textbf{FPS}  & \textbf{consumption (GB)}  \\ \hline \hline 
    Viola-Jones~\cite{Viola2004}      	& 0.6 			& 60.0  		& 0.1    	\\ \hline
    HeadHunter DPM~\cite{DPM}  			& 5.0  			& 1 			& 2.0   	\\ \hline \hline
    SSD\cite{SSD}             			& 45.8   		& 13.3			& 0.7  		\\ \hline
    Faster R-CNN~\cite{FasterRCNN}    	& 223.9  		& 5.8			& 2.1  		\\ \hline
    R-FCN 50~\cite{RFCN}       			& 132.1   		& 6.0			& 2.4 		\\ \hline
    R-FCN 101~\cite{RFCN}       		& 186.6  		& 4.7			& 3.1 		\\ \hline
    PVANET~\cite{PVA}   				& 40.1  		& 9.0			& 2.6  		\\ \hline  
    Local RCNN~\cite{Vu}   				& 1206.8  		& 0.5			& 2.1  		\\ \hline
    Yolo 9000~\cite{Yolo9000}   		& 34.90  		& 19.2			& 2.1  		\\ \hline
\end{tabular}
 \label{Time and memory complexity}
\end{table}

\begin{figure*}[th!]
    \centering
    \includegraphics[width=7.6cm]{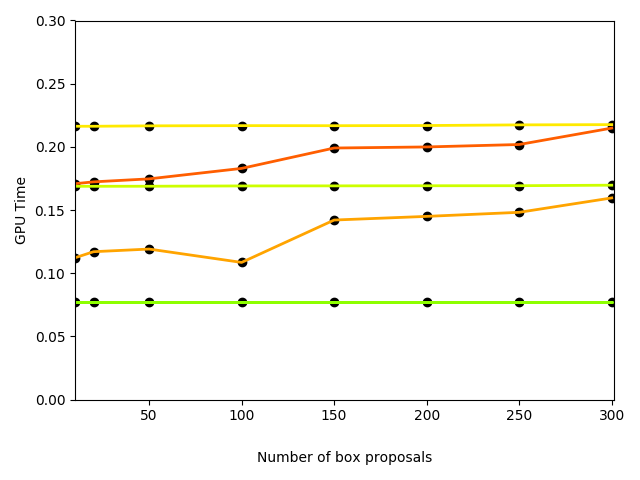}
    \includegraphics[width=7.6cm]{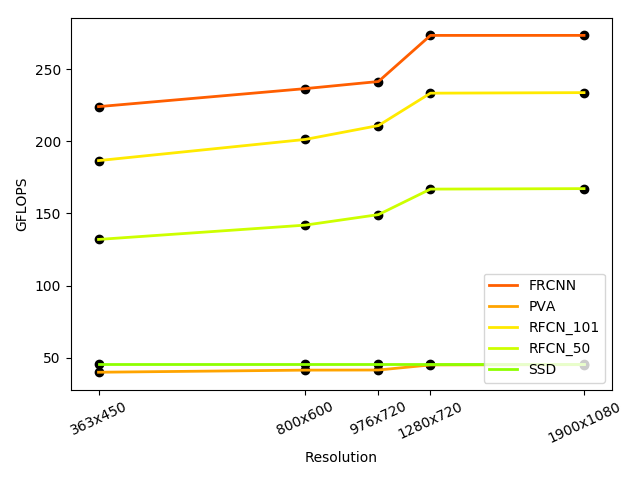}
    \caption{(Right) Average GPU processing time per image in the FDDB dataset as a function of the number of bounding box proposals. (Left) Average number of GFLOPS as a function of image resolution.}
    \label{Box proposals vs GPU Time}
\end{figure*}

Figure \ref{Box proposals vs GPU Time} (Left) shows average GPU time per frame versus the number of bounding box proposals for the CNN detectors on FDDB data. Results indicate that, although the Faster R-CNN and R-FCN architecture consume the highest GPU time per frame, the time complexity of PVANET and Faster R-CNN grows with the number of bounding box proposals. Algorithms with the lowest computational cost, in particular SSD, are most suitable for real-time applications. Figure \ref{Box proposals vs GPU Time} (Right) shows the number of FLOPS according to image resolution. From this figure, we observe that the number of GFLOPS grows on the number of pixels. The number of FLOPS per bit does not grow noticeably for PVANET and SSD. Indeed, SSD resizes images and PVANET uses a concatenation of down-sampled feature map (Hyper Feature) to use the RPN, instead of using RPN directly on a feature map.

Let us consider the application case where a real-time video surveillance system must detect faces or heads in video frames for 10 or more live video feeds, and detect up to 10 different individuals in a single field of view using a computer. Assumed that the surveillance cameras will capture frames at a rate of 20-30 fps and at a SVGA resolution (800x600 pixels) or higher. Implementations must therefore be computationally efficient and scale well to a growing number of cameras, frame rate, resolution and clutter in the scene.

Given the same computer (Intel Xeon CPU E3-1270 with NVIDIA M4000 GPU acceleration card) that was used to produce the results in Table \ref{Time and memory complexity}, it is possible to process multiple video streams at the same time without affecting a detector's algorithmic performance by increasing the batch size during operations (test phase). Since, the CNN detector only uses the GPU, and other operations like encoding and decoding is handled on the CPU, the time complexity would not be heavily impacted. CNN architectures like SDD and PVANET are potentially able to process multiple streams in a same video surveillance system, while sustaining real-time performance. Currently, SSD a requires a comparatively low computational complexity on 450 x 387 pixel images. However, it would appear challenging to achieve real-time performance on HD images (720p, 1080p). The current processing time of SSD is 12.5 fps on 720p images, and 10.3 fps on 1080p images.

%%%%%%%%%%%%%
\section{Conclusion}
%%%%%%%%%%%%%
Computational complexity is an important problem for practical CNN application. The complexity of modern CNN architectures that are suitable for robust face and head detection is high, especially for real-time video surveillance applications. This paper presents a comparison of several single pass and region-based CNN architectures according to accuracy and complexity for face and head detection on public data sets. They are compared empirically to baseline detectors using images from several challenging public datasets. Results shows that while these detectors provide state-of-the-art accuracy, some of the models are not currently suitable for deployment in many real-time video surveillance applications, especially with high definition frames. The SSD and PVANET architectures are potentially capable of real-time application on non-HD images. For head detection, the level of accuracy remains a limitation for real-world environments. Future research should include reducing the complexity of CNN architectures using sparse or perforated convolution, binary CNNs, or other techniques that affect a better trade-off between accuracy and complexity. It should also include evaluating these detectors on more challenging video surveillance datasets.

%%%%% REFS %%%%%%%
\bibliographystyle{abbrv}
\bibliography{refFD.bib}

\begin{thebibliography}{10}

\bibitem{Saman2017}
S.~Bashbaghi, E.~Granger, G.~A. Bilodeau, and R.~Sabourin.
\newblock Dynamic ensembles of exemplar-svms for still-to-video face
  recognition.
\newblock {\em Pattern Recognition}, 6:61--81, 2017.

\bibitem{Bengio12}
Y.~Bengio, A.~C. Courville, and P.~Vincent.
\newblock Unsupervised feature learning and deep learning: {A} review and new
  perspectives.
\newblock {\em CoRR}, abs/1206.5538, 2012.

\bibitem{RFCN}
J.~Dai, Y.~Li, K.~He, and J.~Sun.
\newblock {R-FCN:} object detection via region-based fully convolutional
  networks.
\newblock {\em CoRR}, abs/1605.06409, 2016.

\bibitem{SpeedAcc}
J.~H. et~al.
\newblock Speed/accuracy trade-offs for modern convolutional object detectors.
\newblock {\em CoRR}, abs/1611.10012, 2016.

\bibitem{FasterRCNN}
S.~R. et~al.
\newblock Faster {R-CNN:} towards real-time object detection with region
  proposal networks.
\newblock {\em CoRR}, abs/1506.01497, 2015.

\bibitem{SSD}
W.~L. et~al.
\newblock {SSD:} single shot multibox detector.
\newblock {\em CoRR}, abs/1512.02325, 2015.

\bibitem{PascalVoc}
M.~Everingham, L.~Van~Gool, C.~K.~I. Williams, J.~Winn, and A.~Zisserman.
\newblock The pascal visual object classes (voc) challenge.
\newblock {\em Int'l Journal of Computer Vision}, 2010.

\bibitem{originalDPM}
P.~F. Felzenszwalb, R.~B. Girshick, D.~McAllester, and D.~Ramanan.
\newblock Object detection with discriminatively trained part-based models.
\newblock {\em IEEE Trans. on PAMI}, 32(9):1627--1645, 2010.

\bibitem{FastRCNN}
R.~Girshick.
\newblock Fast {R-CNN}.
\newblock {\em CoRR}, abs/1504.08083, 2015.

\bibitem{RCNN}
R.~Girshick, J.~Donahue, T.~Darrell, and J.~Malik.
\newblock Region-based convolutional networks for accurate object detection and
  segmentation.
\newblock {\em IEEE Trans. on PAMI}, 38(1).

\bibitem{ResNet}
K.~He, X.~Zhang, S.~Ren, and J.~Sun.
\newblock Deep residual learning for image recognition.
\newblock {\em CoRR}, abs/1512.03385, 2015.

\bibitem{FDDB}
V.~Jain and E.~Learned-Miller.
\newblock Fddb: A benchmark for face detection in unconstrained settings.
\newblock Technical Report UM-CS-2010-009, University of Massachusetts,
  Amherst, 2010.

\bibitem{PVA}
K.~Kim, Y.~Cheon, S.~Hong, B.~Roh, and M.~Park.
\newblock {PVANET:} deep but lightweight neural networks for real-time object
  detection.
\newblock {\em CoRR}, abs/1608.08021, 2016.

\bibitem{HyperNet}
T.~Kong, A.~Yao, Y.~Chen, and F.~Sun.
\newblock Hypernet: Towards accurate region proposal generation and joint
  object detection.
\newblock {\em CoRR}, abs/1604.00600, 2016.

\bibitem{Oquab}
M.~Oquab, L.~Bottou, I.~Laptev, and J.~Sivic.
\newblock Learning and transferring mid-level image representations using
  convolutional neural networks.
\newblock In {\em IEEE Conference on CVPR}, pages 1717--1724, 2014.

\bibitem{Yolo9000}
J.~Redmon and A.~Farhadi.
\newblock {YOLO9000:} better, faster, stronger.
\newblock {\em CoRR}, abs/1612.08242, 2016.

\bibitem{VJ-CRF}
X.~Ren.
\newblock Finding people in archive films through tracking.
\newblock In {\em IEEE Conference on CVPR}, 2008.

\bibitem{Viola2004}
P.~Viola and M.~J. Jones.
\newblock Robust real-time face detection.
\newblock {\em Int. J. Comput. Vision}, 57(2):137--154, May 2004.

\bibitem{Vu}
T.~Vu, A.~Osokin, and I.~Laptev.
\newblock Context-aware {CNNs} for person head detection.
\newblock In {\em ICCV}, 2015.

\bibitem{DPM}
J.~Yan, X.~Zhang, Z.~Lei, and S.~Z. Li.
\newblock Real-time high performance deformable model for face detection in the
  wild.

\bibitem{WIDER}
S.~Yang, P.~Luo, C.~C. Loy, and X.~Tang.
\newblock Wider face: A face detection benchmark.
\newblock In {\em IEEE Conference on CVPR}, 2016.

\bibitem{ZAFEIRIOU20151}
S.~Zafeiriou, C.~Zhang, and Z.~Zhang.
\newblock A survey on face detection in the wild: Past, present and future.
\newblock {\em Computer Vision and Image Understanding}, 138:1 -- 24, 2015.

\end{thebibliography}

\end{document}